# PREDICTING NONLINEAR SEISMIC RESPONSE OF STRUCTURAL BRACES USING MACHINE LEARNING

## (TO BE PRESENTED IMECE2020-24014)

**Elif Ecem Bas[1], Denis Aslangil[2], Mohamed Aly Moustafa[1]**

[1] Department of Civil Engineering, University of Nevada – Reno, Reno NV 89557
[2]Computational Physics and Methods, CCS-2, Los Alamos National Laboratory – Los Alamos, NM 87545

## ABSTRACT

*Numerical modeling of different structural materials that have highly nonlinear behaviors has always been a challenging problem in engineering disciplines. Experimental data is commonly used to characterize this behavior. This study aims to improve the modeling capabilities by using state of the art Machine Learning techniques, and attempts to answer several scientific questions: (i) Which ML algorithm is capable and is more efficient to learn such a complex and nonlinear problem? (ii) Is it possible to artificially reproduce structural brace seismic behavior that can represent real physics? (iii) How can our findings be extended to the different engineering problems that are driven by similar nonlinear dynamics? To answer these questions, the presented methods are validated by using experimental brace data. The paper shows that after proper data preparation, the long-short term memory (LSTM) method is highly capable of capturing the nonlinear behavior of braces. Additionally, the effects of tuning the hyperparameters on the models, such as layer numbers, neuron numbers, and the activation functions, are presented. Finally, the ability to learn nonlinear dynamics by using deep neural network algorithms and their advantages are briefly discussed.*

## NOMENCLATURE

| | |
|---|---|
| DNN | Deep neural network |
| HS | Hybrid simulation |
| LSTM | Long-short term memory |
| ML | Machine learning |
| NRMSE | Normalized root mean square error |
| PIML | Physics informed machine learning |
| RNN | Recurrent neural network |
| RTHS | Real-time hybrid simulation |
| STS | Structural Test System |

## 1. INTRODUCTION

Braces are widely used in structural systems to provide lateral load resistance. Thus, the structural response of the concentrically braced frames under extreme events (such as earthquakes, winds, blasts) strongly depends on its brace behavior [1]. In scientific literature, a significant amount of studies has investigated the brace behavior both numerically and experimentally [1-5]. However, due to the highly nonlinear and non-uniform characteristics of the braces, which also depends on the loading histories, it is still an open question on how to model their dynamic behaviors accurately [6,7]. A robust model has to capture bucking, deterioration, and failure due to low-cycle fatigue. These effects strongly depend on the nonlinear behavior of the braces, which is still not well understood in some cases, e.g. under longer duration earthquakes, and thus difficult to numerically model [6].

Recently, there have been efforts to improve the numerical model capabilities by benefiting from the high-fidelity experimental data. To this end, machine learning (ML) provides new opportunities for scientists to answer long-standing questions and serves promising solutions in many dynamic engineering problems [8-9]. It is a developing field in structural engineering where a wide range of applications are considered, such as predicting the global structural response under dynamic excitations [10]. In this study, we aim to further improve the analytical models by using the advantages of physics-informed ML. For this purpose, small-scale brace responses, which are tested under increasing scale cyclic loading, is used as the training dataset. As a deep neural network algorithm, deep long-short term memory (LSTM) networks that could predict the time-series data sequences are used to predict the dynamic responses of the braces. Several models are generated by changing the neuron and number of layers as well as time steps (lookback) to tune the hyperparameters and make accurate predictions.

In this study, we propose an approach where we predict the nonlinear behavior of the structural braces using a data-driven model. The aim is to be able to model the nonlinear response behavior by using ML algorithms. The training dataset of the ML algorithm is an experimental dataset which is obtained from cyclic loading tests and explained in detail. LSTM networks are selected to be used as an ML algorithm. It is shown that the LSTM models exhibit a great potential to model the nonlinear material behavior due to having sequence to sequence input-output relationship capabilities. Moreover, this modeling assumption has great potential to significantly decrease the number of necessary experiments for characterizing braces with different materials.

[1] Contact author: basel@nevada.unr.edu

The structure of this paper is as follows: Section 2 describes the experimental setup and methods. Section 3 presents the results, and Section 4 summarizes our main findings.

## 2. EXPERIMENTAL SETUP AND TRAINING DATASET

### 2.1 Experimental setup

In this study, a small-scale load frame that is part of a recently developed hybrid simulation (HS) setup, but could still be used independently, at the University of Nevada, Reno, is used for obtaining the experimental dataset (see Figure 1).

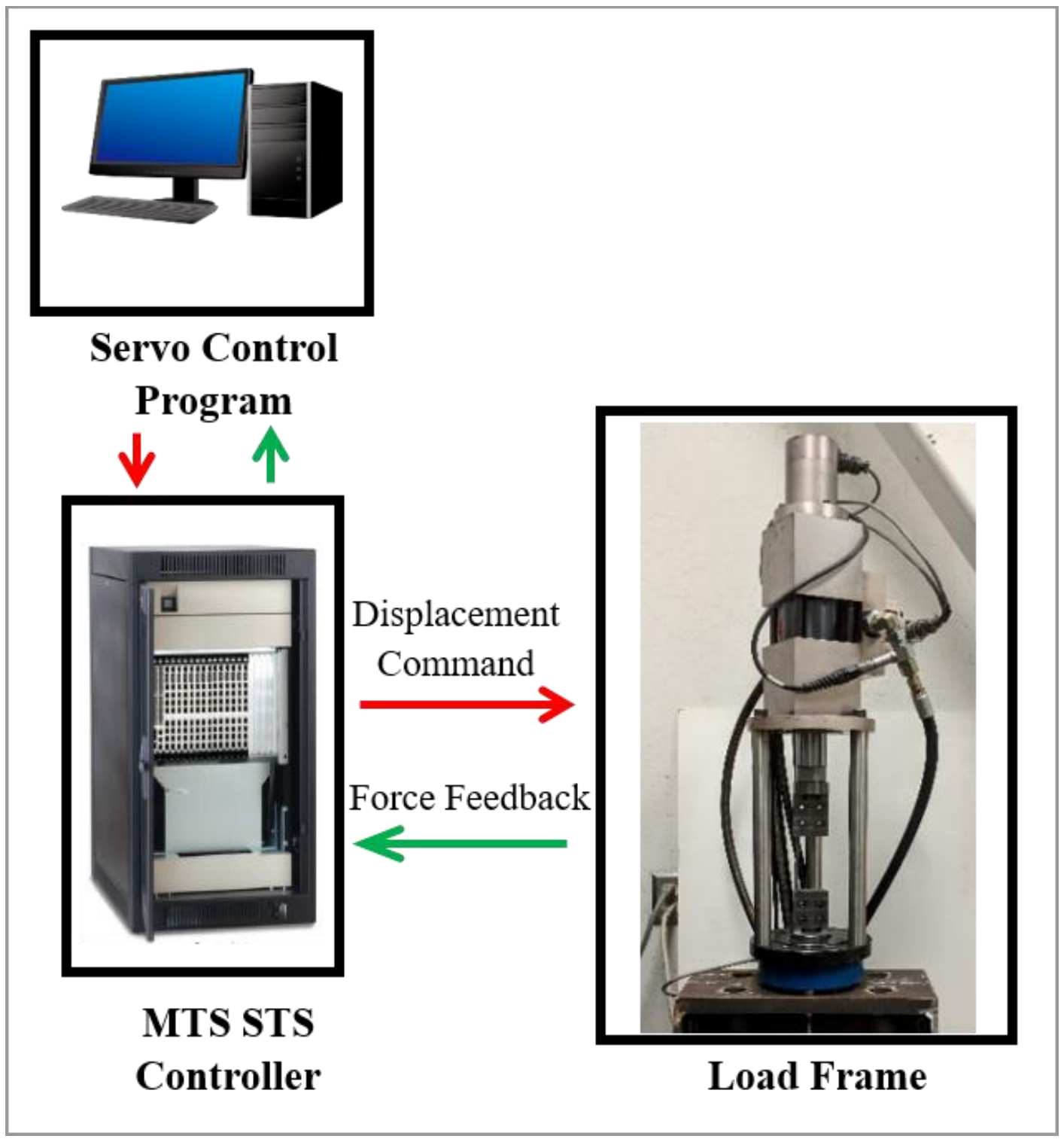

**FIGURE 1:** The representative scheme of the experimental setup.

This load frame is mainly used as an experimental setup for the hybrid simulation system; however, it can be controlled individually to do material testing. Here, we only introduce the parts that are used for material testing (see Figure 1). However, the reader is referred to Bas et al. [6] for more details about HS system capabilities and its validation.

The load frame has a dynamic actuator where the maximum load capacity is 31.14 kN (7 kips) and has ±25.4 mm (±1 in) stroke. The actuator can achieve the peak velocity of 338.84 mm/sec (13.34 in/sec) at no load. In addition, the system is supported with an isolated hydraulic power supply system which provides an 8.71 l/min (2.3 gpm) pumping capacity and the reservoir capacity of oil volume is 56.78 lt (15 gallons). MTS Structural Test System (STS) 493 Hardware Controller manages the motion of the actuator where the Servo Controller Program is used to access and modify the control properties through the Servo Controller Program.

### 2.2 Specimens and Loading Protocol

In structural systems, braced frames are widely used, especially in seismically active areas to provide lateral load resisting systems. Usually, the brace members are made of structural steel and ideally buckle under axial compression loading and yield under axial tension loading [6]. Table 1 presents the material properties of the two different dog-bone specimens with different material properties where Specimen A and Specimen B are steel and aluminum, respectively. These specimens are used to generate two distinct experimental datasets. Both specimens are tested under increasing scale cyclic loading until failure to capture buckling, deterioration, and failure due to low-cycle fatigue induced rupture.

| Properties | Specimen A | Specimen B | Specimen dimensions [mm] |
|---|---|---|---|
| Section cross section [mm x mm] | 6.35x6.35 | 6.35x12.7 | 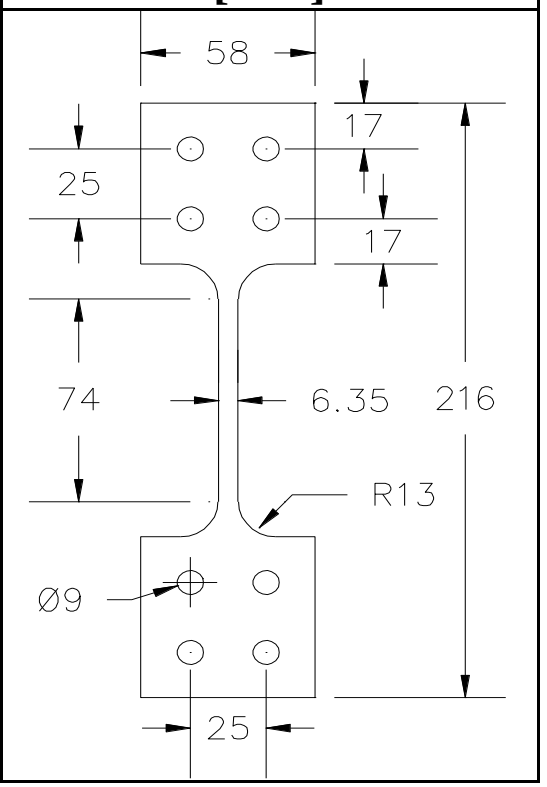 |
| Yield Strength[MPa] | 413 | 145 | |
| Fabrication | Cold Worked | Cold Worked | |
| Temper Rating | Hardened | H14 (1/2 Hard) | |
| Hardness | Rockwell 80 (Medium) | Brinnel 40 (Soft) | |

**TABLE 1:** Properties of Specimen A (steel) and Specimen B (aluminum).

The Modified SAC loading protocol that is adopted from Lumpkin [5] is used as the loading for the experiments. Displacement controlled cyclic loads are applied until the brace failure. Table 2 and Figure 2 present the adopted loading protocol where loading amplitudes are expressed in terms of yielding displacement of the tested brace.

| Peak | ±0.5Δy | | ±0.75Δy | ±1Δy | ±1.5Δy | ±2Δy | ±3Δy |
|---|---|---|---|---|---|---|---|
| Cycle | 1,2 | | 3,4 | 5,6 | 7,8 | 9,10 | 11,12 |

| Peak | ±4Δy | ±5Δy | ±6Δy | ±7Δy | ±8Δy | ±9Δy | ±10Δy |
|---|---|---|---|---|---|---|---|
| Cycle | 13,14 | 15,16 | 17,18 | 19,20 | 21,22 | 23,24 | 25,26 |

**TABLE 2:** Loading Protocol for the cyclic test.

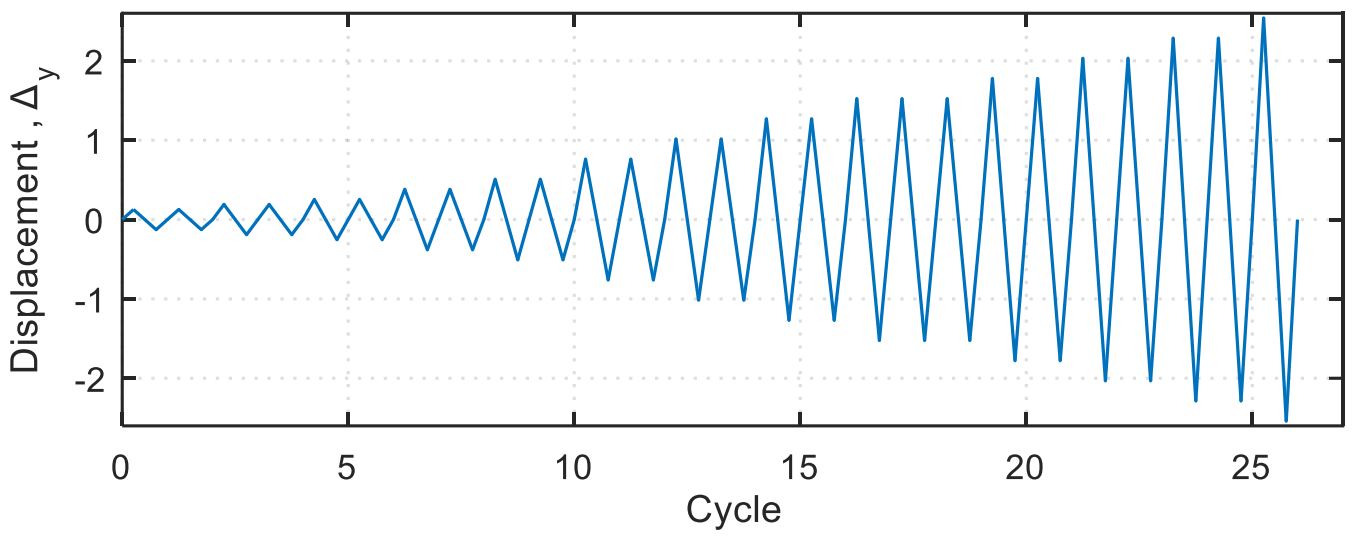

**FIGURE 2:** Cyclic-Loading Protocol

### 2.3 Experimental Results

Two experiments are conducted applying the previously defined displacement protocols to the two different specimens. Figures 3(a) and 4(a) show the applied displacement time histories and as it is seen the displacement magnitude is increasing in time and they include both tension and compression displacements. The generated response in terms of the force (kips) of the two braces to the displacement profiles can be seen in figures 3b and 4b. It is clear from the figures that the behavior of the braces is highly nonlinear as, initially, there is a positive correlation between the applied displacement and the force where the force is increasing upon an increase in the displacement. However, later, the larger displacements start to generate smaller forces so the correlation between the displacement and force becomes negative. Capturing this nonlinear behavior is a great challenge for the numerical methods, and it requires the reforming of significantly large amounts of long experiments for each different specimen. In addition, Figures 3(c) and 4(c) present the brace hysteresis where the braces experience buckling and low-cycle fatigue induced rupture which is hard to model accurately with the current finite element modeling techniques. Since the overall structural behaviors of braced frames are highly dependent on the hysteretic response of the braces, it is very important to model this behavior numerically. The presented experimental datasets are used for training and testing the ML algorithms that are discussed in detail in the next section.

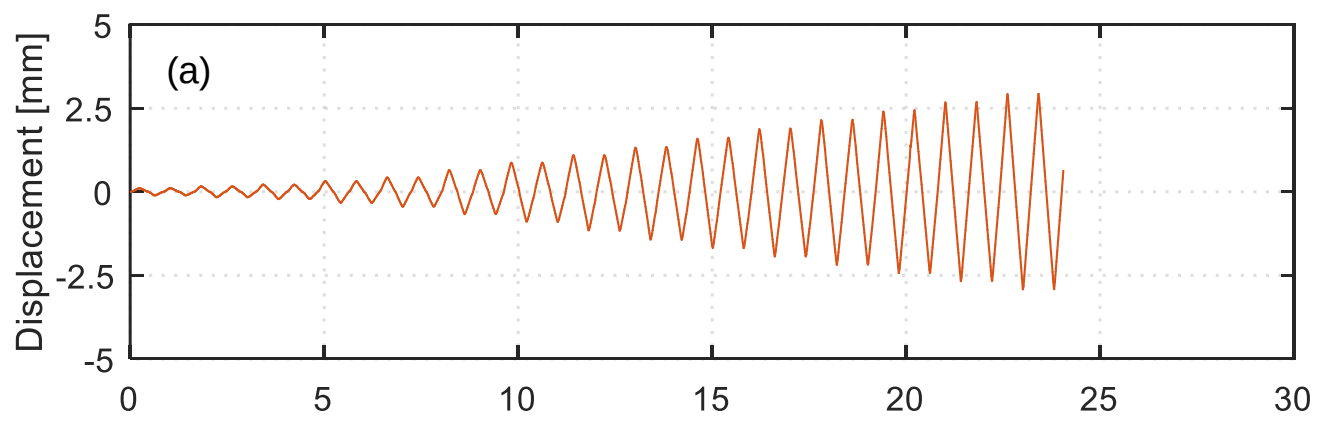

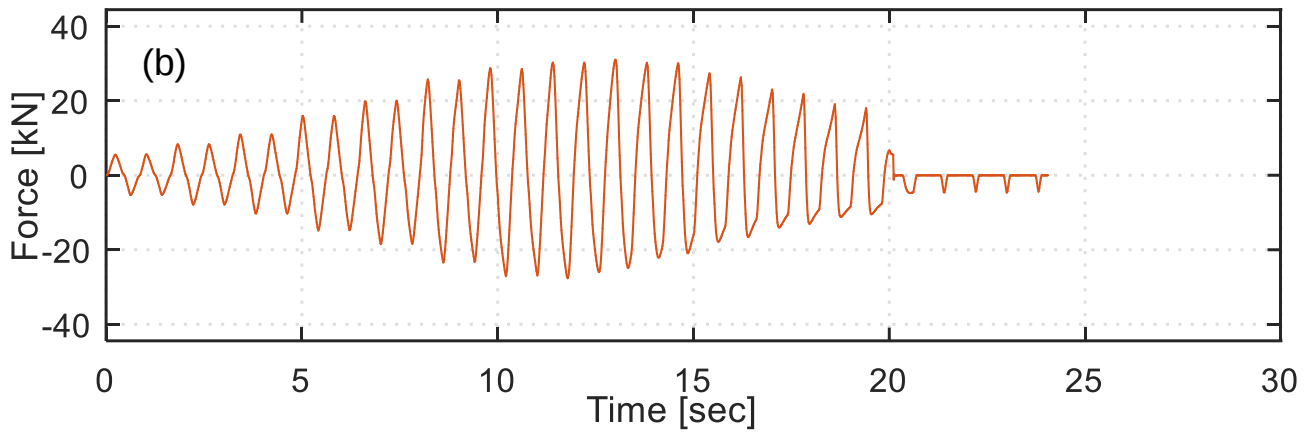

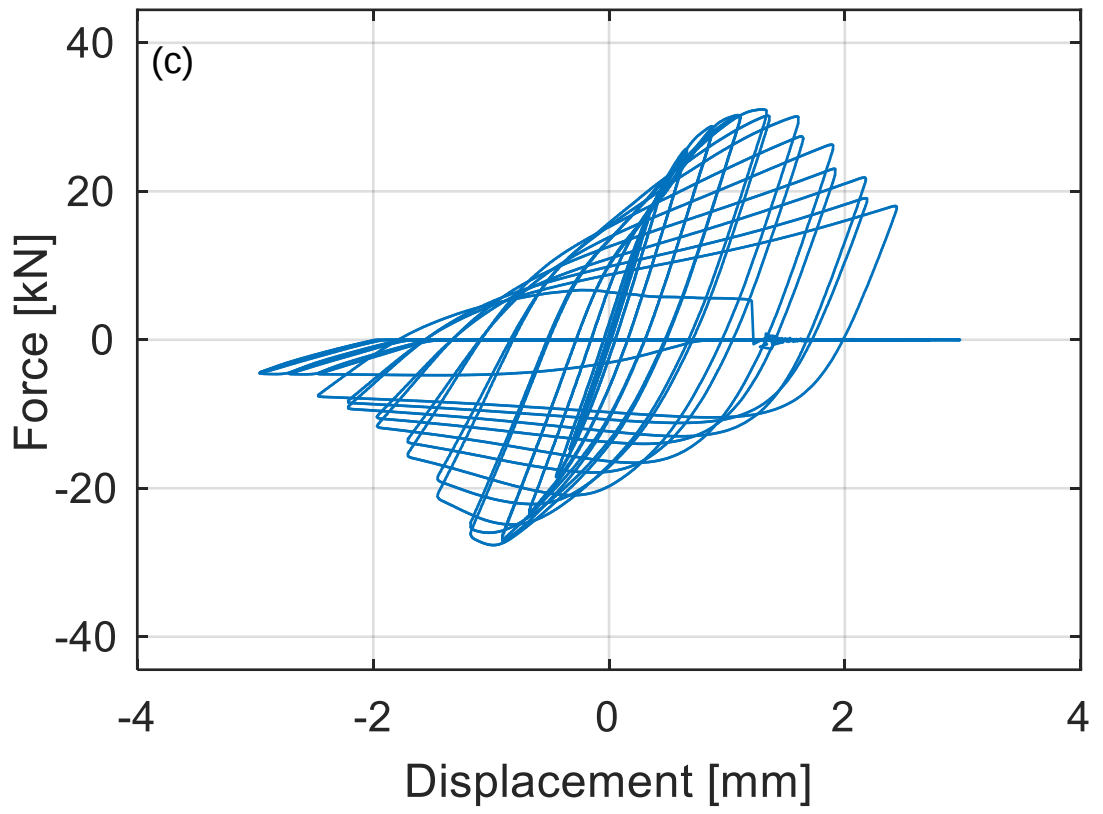

**FIGURE 3:** Specimen A: (a) Displacement time history, (b) Force time history, (c) Brace hysteresis

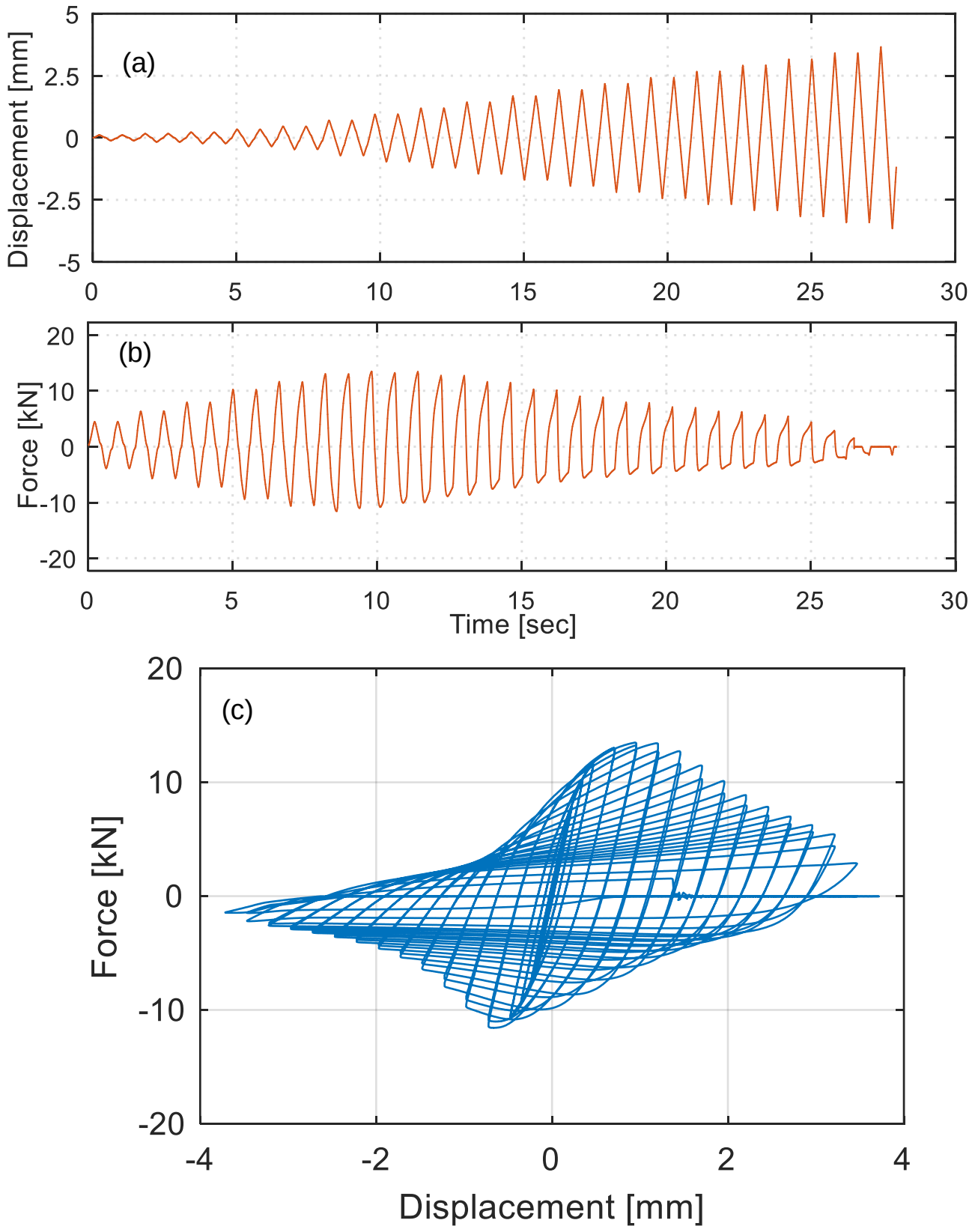

**FIGURE 4:** Specimen B: (a) Displacement time history, (b) Force time history, (c) Brace hysteresis.

## 3. LSTM NETWORK FOR MODELING

Recurrent neural networks (RNN) are widely used in forecasting the time series. The great advantage of the RNN algorithms is the backward connection points, which allows the layers to receive both inputs as well as their own outputs from the previous time step. However, the method itself comes with two main disadvantages: (i) having unstable gradients and (ii) utilizing a very limited short-term memory. In order to tackle these problems, LSTM cells are recommended as they perform better, converge rapidly, and detect the long-time dependencies in the data [12].

The LSTM networks are trained with multiple datasets, and as in other time series prediction models, the input sequences are formatted in three-dimensional arrays. The first dimension is the batch size, which consists of independent datasets; the second dimension is the time steps that can also be defined as *lookback*, and the third dimension is the size of the input dimension. Dimensionality is equal to unity for the problems where the input has only one variable and can be larger than one for the multivariate time series. Lookback parameter is one of the key parameters that makes LSTMs more reliable since it allows the algorithm to use "lookback" in the past time steps. Hence, it extends the used information that leads to better predictions for the next time steps, although it slightly increases the memory requirements.

The LSTM cells include mainly two vectors, namely $\mathbf{h}_{(t)}$ and $\mathbf{c}_{(t)}$, where $\mathbf{h}_{(t)}$ represents the short-term state and $\mathbf{c}_{(t)}$ represents the long-term state. For each time step, the current input vector $\mathbf{x}_{(t)}$ and the previous time step input $\mathbf{h}_{(t-1)}$ are fed into four fully connected layers which all serve different purposes. The $\mathbf{g}_{(t)}$ state's role is to analyze the current inputs and the previous short-term state, and this state's most important parts of being stored in the long-term state, $\mathbf{c}_{(t)}$, where the rest is dropped. The other three layers are the gate controllers. $\mathbf{f}_{(t)}$ is the forget gates which control the erased parts of the long-term state. Input gate $\mathbf{i}_{(t)}$ decides which parts of the $\mathbf{g}_{(t)}$ state should proceed in the long-term state. And the output state $\mathbf{o}_{(t)}$ is where the decision of long-term state parts should be the output of the current time step for both short term state $\mathbf{h}_{(t)}$ and the output state. The LSTM computations are given in Equation (1), and a typical LSTM cell is shown in Figure 5. In the equations, $\boldsymbol{W}_x$s are the weight matrices of each layer for their connection to the input vector $\mathbf{x}_{(t)}$, where the $\boldsymbol{W}_h$s are the weight matrices of each layer for their connection to the previous short-term state. Bias terms are represented with $\boldsymbol{b}$s in each layer [11].

$$\begin{aligned}
\boldsymbol{i}_{(t)} &= \sigma\left(\boldsymbol{W}_{xi}{}^{T}\boldsymbol{x}_{(t)} + \boldsymbol{W}_{hi}{}^{T}\boldsymbol{h}_{(t-1)} + \boldsymbol{b}_{i}\right)\\
\boldsymbol{f}_{(t)} &= \sigma\left(\boldsymbol{W}_{xf}{}^{T}\boldsymbol{x}_{(t)} + \boldsymbol{W}_{hf}{}^{T}\boldsymbol{h}_{(t-1)} + \boldsymbol{b}_{f}\right)\\
\boldsymbol{o}_{(t)} &= \sigma\left(\boldsymbol{W}_{xo}{}^{T}\boldsymbol{x}_{(t)} + \boldsymbol{W}_{ho}{}^{T}\boldsymbol{h}_{(t-1)} + \boldsymbol{b}_{o}\right)\\
\boldsymbol{g}_{(t)} &= tanh\left(\boldsymbol{W}_{xg}{}^{T}\boldsymbol{x}_{(t)} + \boldsymbol{W}_{hg}{}^{T}\boldsymbol{h}_{(t-1)} + \boldsymbol{b}_{g}\right)\\
\boldsymbol{c}_{(t)} &= \boldsymbol{f}_{(t)} \otimes \boldsymbol{c}_{(t-1)} + \boldsymbol{i}_{(t)} \otimes \boldsymbol{g}_{(t)}\\
\boldsymbol{y}_{(t)} &= \boldsymbol{h}_{(t)} = \boldsymbol{o}_{(t)} \otimes tanh\left(\boldsymbol{c}_{(t)}\right)
\end{aligned} \qquad (1)$$

As explained above, an LSTM cell can observe the importance of the input, can remember the longer history of the time series storing in the long-term state, and can store longer information as long as it is needed and remove the information whenever it is unnecessary. The advantages of the LSTM make it suitable to capture the long-term patterns in time series even if the problem is highly nonlinear [12]. As a result, in this paper, deep LSTM networks are used for multiple hidden layers that include LSTM layers and fully connected layers. The hyperparameters that are tuned here are neuron networks, hidden layers, and lookback numbers and they are discussed next.

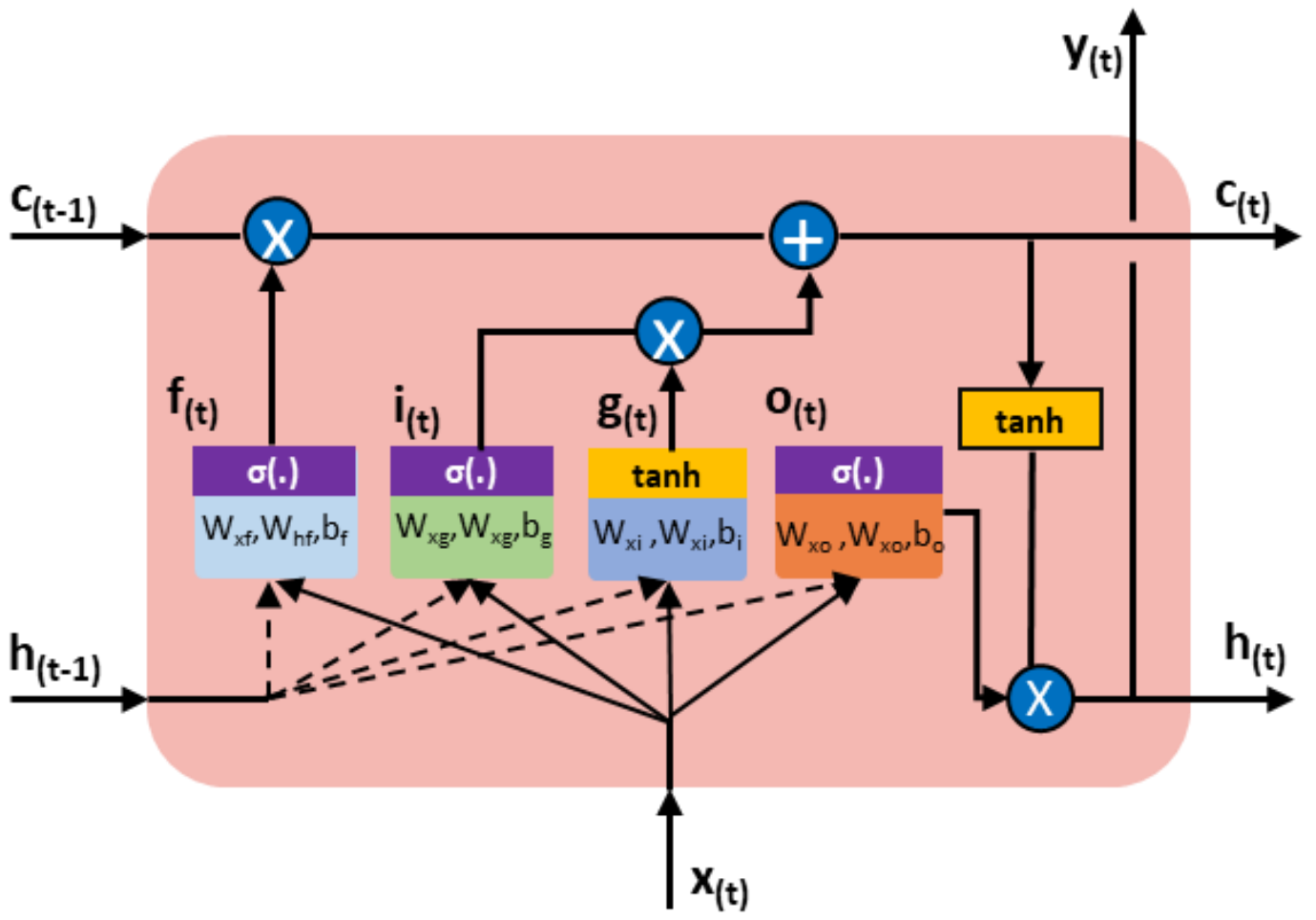

**FIGURE 5:** Typical LSTM cell

### 3.1 Methodology for LSTM

Several LSTM models are prepared to tune the hyperparameters of the network in order to obtain a proper model to predict the nonlinear brace responses. As mentioned earlier, the dataset to train the model is obtained from the brace experiments. The displacement time histories are used as input where the output is the force of the brace which is the prediction of the model.

Table 3 shows the hyperparameters of the LSTM models that are used for both materials of this study. Mainly three models are developed and named with respect to their neuron numbers, where Model 1 has 5, Model 2 has 20, and Model 3 has 40 neurons. For Model 1 and Model 2, the hidden layers are set to be 5 where the lookback parameter is 30 for both models. Five different versions of Model 3 are generated by changing the hidden layers in the range of 5 to 20, where lookback parameters are also changed from 10 to 40. For both materials, 50% of the experimental data is used as the training dataset, while the rest is used as the test dataset. The model performances are evaluated by calculating the normalized root mean square errors (NRMSE).

|  | Neuron # | Hidden layer # | Lookback |
|---|---|---|---|
| Model 1 | 5 | 5 | 30 |
| Model 2 | 20 | 5 | 30 |
| **Model 3a** | **40** | **5** | **30** |
| Model 3b | 40 | 10 | 30 |
| Model 3c | 40 | 20 | 30 |
| Model 3d | 40 | 5 | 10 |
| Model 3e | 40 | 5 | 40 |

**TABLE 3:** LSTM hyperparameters.

### 3.3 Comparison of the experimental and predicted results

As mentioned above, the overall experimental data is divided into two halves, where the first half represents the training dataset and the other half is the test dataset. The input of the deep LSTM network is designed to be the brace displacement which predicts the brace axial force as the output of the ML model. Here, only the results from Model 3a are shown, which gives the most accurate results. The NRMSE values for both material modeling predictions are calculated as 18% and 14% for Specimen A and Specimen B, respectively.

The comparison between the true force values and the predicted force values are shown in Figure 5 and Figure 6 for Specimen A and B, respectively. As can be seen from the figures, the strength degradation can be predicted successfully by the deep LSTM model, even with the limited portion of the data used in training. In other words, although the LSTM algorithm is trained by the portion where there is a positive correlation between the displacement and force, the algorithm is able to capture the negative correlation between these two metrics.

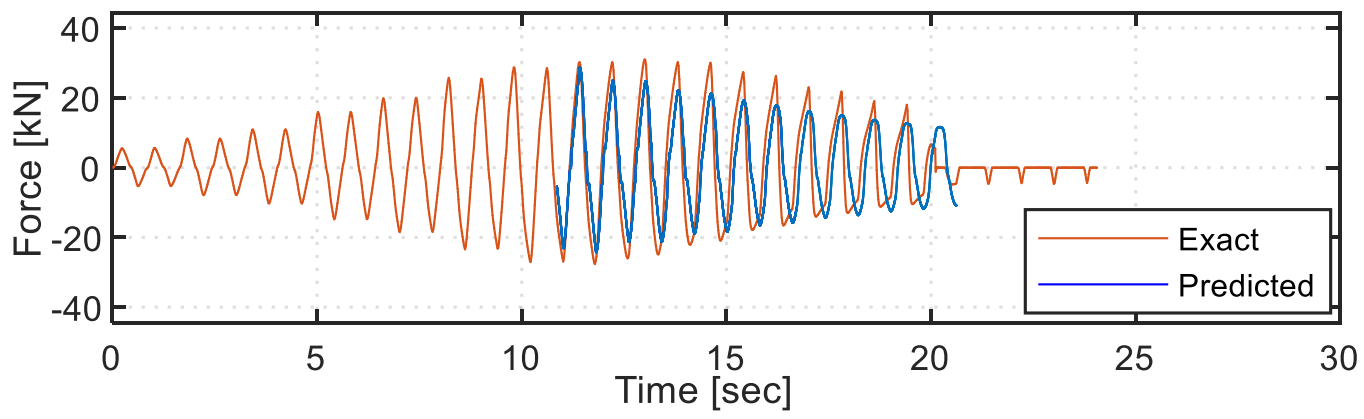

**FIGURE 5:** Force time history comparison for Specimen A

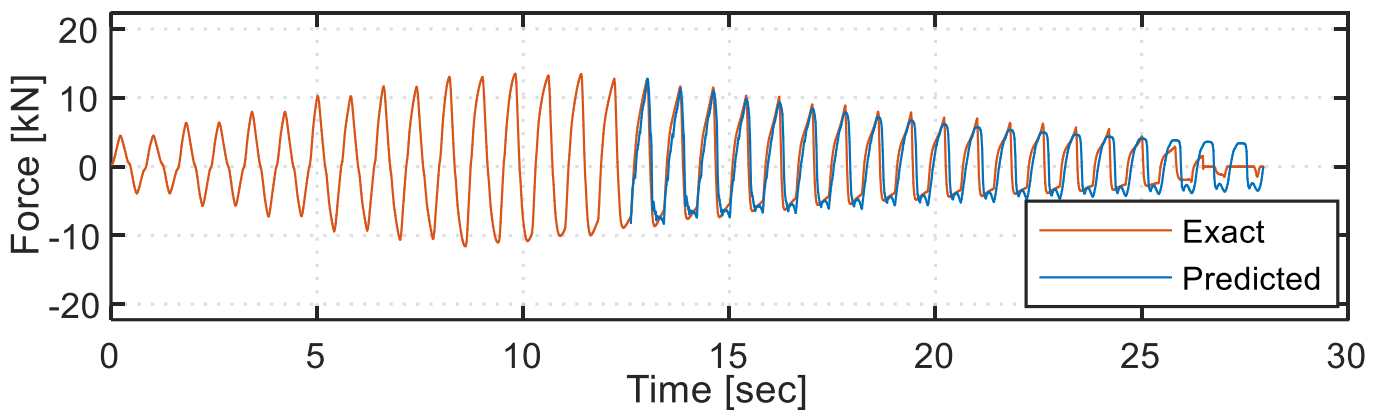

**FIGURE 6:** Force time history comparison for Specimen B

## 4. SUMMARY AND CONCLUSIONS

In this study, experimental cyclic loading results of the steel and aluminum small-scale braces, which were obtained from the experimental setup at the University of Nevada, Reno, are used to explore the suitability of ML techniques on modeling nonlinear behaviors. The key findings can be summarized as:

- Deep LSTM networks can be used to forecast the nonlinear seismic brace behavior due to their success in capturing time-series using long- and short-term memories.
- Similar hyperparameter values are found to be optimal for modeling two different materials with an LSTM technique.
- By proper data preparation, the LSTM models are able to capture the nonlinear behavior of the materials. However, increasing the number of training data, i.e. using more experimental data, different loading protocols, etc., beyond what has been used herein is needed to further improve the prediction accuracy of the model.

For future directions, it would be interesting to run new experiments and to compare LSTM predictions with different loading protocols such as for earthquakes, wind, and so on [6]. The found capability of the LSTM technique to capture nonlinear behaviors is promising to extend its applicability to the different nonlinear problems from other engineering areas, such as for turbulent mixing problems occuring in atmospheric, oceanographic, and astrophysical flows [13, 14].